\documentclass[runningheads]{llncs}
\usepackage[T1]{fontenc}
\usepackage{hyperref}
\usepackage{geometry}
\usepackage{graphicx}
\usepackage{amsmath}
\usepackage[normalem]{ulem}

\usepackage{xcolor} 
\usepackage{tikz}
\usetikzlibrary{shapes,arrows.meta,positioning,calc,trees,decorations.markings,shadows}
\usepackage{subcaption}
\usepackage{ulem}
\usepackage{caption}
\usepackage{listings}
\usepackage[most]{tcolorbox}
\tcbset{
    academicexample/.style={
        colback=gray!10,
        colframe=gray!70,
        sharp corners,
        boxrule=0.8pt,
        fontupper=\scriptsize,
        left=8pt, right=8pt,
        top=6pt, bottom=6pt,
        width=\linewidth
    }
}
\usepackage{float}

\definecolor{maltPink}{HTML}{F9E7E7}
\definecolor{maltGreenFill}{HTML}{E8F4F1}
\definecolor{topdownBlack}{HTML}{333333}
\definecolor{bottomupChaos}{HTML}{E74C3C}
\definecolor{orphanOrange}{HTML}{E67E22}
\definecolor{wikidataBlue}{HTML}{005689}
\definecolor{preprocessorRed}{HTML}{E74C3C}
\definecolor{preprocessorFill}{HTML}{FADBD8}
\definecolor{maltGreen}{HTML}{1A9073}

\title{An Agentic Hybrid Top-Down and Bottom-Up Approach to Knowledge Graph Generation}
\author{Emma Jouffroy\orcidID{0009-0003-2930-0330} \and
Warren Jouanneau\orcidID{0000-0003-4973-2416} \and
Marc Palyart\orcidID{0000-0002-8808-0492}}
\authorrunning{E. Jouffroy et al.}
\institute{Malt, Paris, France}
\date{\today}

\begin{document}

\maketitle
\begin{abstract}
Organizing thousands of unstandardized, multilingual expertise declarations is a persistent challenge for Human Resources (HR) platforms, directly impacting downstream tasks like accurate talent matching. To address this, we propose a hybrid knowledge graph generation pipeline that grounds a Large Language Model (LLM) in the Wikidata multilingual Knowledge Graph (KG) while employing an agentic reflexion pattern to synthesize emerging concepts and their associated metadata. Unlike rigid top-down methods or fragmented bottom-up approaches, our system anchors recognized concepts to stable Knowledge Graph entities while dynamically creating new nodes and relational metadata for unrecognized skills. Executed across five stages, entity reconciliation, multilingual canonicalization, active curation, deduplication, and the iterative recovery of unmapped concepts, the system autonomously adapts to rapidly evolving, noisy skill mentions across five European languages. Ultimately, this pipeline provides a highly scalable, explicable, and self-healing framework for generating a comprehensive skills knowledge graph, from which a structured taxonomy is derived, using unstructured, noisy text.  
\keywords{Large Language Models \and Knowledge Graphs \and Entity Reconciliation \and Responsible HR AI.}
\end{abstract}

\section{Introduction}
Comprehensive, structured skills knowledge graphs and their derivative taxonomies serve as a critical foundation for HR applications, such as standardizing job requirements and facilitating talent matching. However, maintaining these structures is a persistent challenge for talent marketplaces like Malt, the European freelancer marketplace. Free-text expertise declarations frequently lead to multilingual fragmentation, niche jargon, compound skills, and rapid temporal drift as occupations and work practices evolve. Although Large Language Models (LLMs) overcome the scalability issues of hand-crafted taxonomies, limitations remain, such as a susceptibility to hallucinated skills, formatting instability, and a lack of grounding in verifiable metadata \cite{climb_citation,tnt_llm_citation}.

\begin{figure}[htbp]
    \centering
    \resizebox{0.75\linewidth}{!}{%
        \begin{tikzpicture}[
    node distance=1.2cm and 0.6cm,
    baseBox/.style={rectangle, rounded corners=6pt, draw=none, inner sep=6pt, align=center, font=\small\bfseries, drop shadow={shadow xshift=1.5pt, shadow yshift=-1.5pt, fill=maltCharcoal!12, opacity=0.4}},
    taxonomyNode/.style={baseBox, fill=maltCream, text=maltCharcoal, text width=2.8cm},
    targetNode/.style={baseBox, fill=maltCharcoal, text=maltCream, text width=3.4cm},
    coralNode/.style={baseBox, fill=maltCoralLight, text=maltCharcoal, text width=3.4cm},
    labelNode/.style={align=center, font=\scriptsize\bfseries, text=maltCharcoal!90, text width=2.0cm, inner sep=2pt},
    coralLabelNode/.style={align=center, font=\scriptsize\bfseries, text=maltCoral, text width=2.2cm, inner sep=2pt},
    trackBanner/.style={align=center, font=\fontseries{b}\selectfont\scriptsize, text width=3.2cm, inner sep=2pt},
    skillDot/.style={circle, fill=maltCharcoal!60, inner sep=0pt, minimum size=2.8mm},
    outlierDot/.style={circle, fill=maltCoral, inner sep=0pt, minimum size=2.8mm},
    edgeLine/.style={-{Latex[length=5pt, width=5pt]}, draw=maltCharcoal, line width=1.2pt},
    failLine/.style={-{Latex[length=5pt, width=5pt]}, draw=maltCoral, dashed, line width=1.2pt},
    loopLine/.style={-{Latex[length=5pt, width=5pt]}, draw=maltCoral, line width=1.5pt}
]

\definecolor{maltCoral}{HTML}{FC5757}       
\definecolor{maltCoralLight}{HTML}{FFF2F2}  
\definecolor{maltCharcoal}{HTML}{2C232E}   
\definecolor{maltCream}{HTML}{FAF8F5}

\begin{scope}[xshift=0cm]
    \node (aRoot) [taxonomyNode] at (0, 3.6) {Core Ontology};
    \node (aPM) [taxonomyNode] at (0, 2.2) {Project\\ Management};
    \draw [edgeLine] (aPM) -- (aRoot);

    \node (dotPM) [skillDot] at (-1.0, 0.6) {};
    \node (dotWebPM) [outlierDot] at (1.0, 0.6) {};
    
    \node (lblPM) [labelNode, below=4pt of dotPM] {Project\\ Management};
    \node (lblWebPM) [labelNode, below=4pt of dotWebPM] {Web Project\\ Management};

    \draw [edgeLine] (dotPM) -- (dotPM |- aPM.south);
    
    \draw [draw=maltCoral, line width=1.8pt] (0.3, 1.5) -- (1.7, 1.5); 
    \draw [failLine] (dotWebPM) -- (1.0, 1.5);
    \node [text=maltCoral, font=\fontseries{b}\selectfont\scriptsize, anchor=west, text width=1.2cm, align=left] at (1.1, 1.0) {Missing\\ Node};

    \node [font=\normalsize\bfseries, text=maltCharcoal] at (0, -0.8) {(A) Rigid Top-Down};
\end{scope}

\begin{scope}[xshift=7.0cm]
    \node (bCluster1) [taxonomyNode, fill=white, text width=2.4cm, drop shadow={shadow xshift=1.5pt, shadow yshift=-1.5pt, fill=maltCharcoal!8, opacity=0.3}] at (-1.4, 2.2) {PM Cluster};
    \node (bCluster2) [taxonomyNode, fill=white, text width=2.4cm, drop shadow={shadow xshift=1.5pt, shadow yshift=-1.5pt, fill=maltCharcoal!8, opacity=0.3}] at (1.4, 2.2) {Web PM\\ Group};

    \node (bDot1) [skillDot] at (-1.4, 0.6) {};
    \node (bDot2) [skillDot] at (0.0, 0.6) {};
    \node (bDot3) [skillDot] at (1.4, 0.6) {};
    
    \node (lblb1) [labelNode, below=4pt of bDot1] {Project\\ Management};
    \node (lblb2) [labelNode, below=4pt of bDot2] {Web PM};
    \node (lblb3) [labelNode, below=4pt of bDot3] {Excel};

    \draw [edgeLine] (bDot1) -- (bCluster1.south);
    \draw [edgeLine] (bDot2) -- (bCluster1.south);
    \draw [failLine] (bDot3) -- (bCluster2.south);
    
    \draw [<->, draw=maltCoral, dotted, line width=1.5pt] (bDot1) to[bend left=22] (bDot2);
    \node [text=maltCoral, font=\fontseries{b}\selectfont\scriptsize, text width=1.4cm, align=center] at (-0.8, 1.2) {Fragmented};
    \node [text=maltCoral, font=\fontseries{b}\selectfont\scriptsize, anchor=west, text width=1.5cm, align=left] at (0.9, 1.1) {Hallucination};

    \node [font=\normalsize\bfseries, text=maltCharcoal] at (0, -0.8) {(B) Chaotic Bottom-Up};
\end{scope}

\begin{scope}[xshift=15.4cm]
    \node (trackTD) [trackBanner, text=maltCharcoal!60] at (-2.2, 3.6) {TOP-DOWN\\ GROUNDING};
    \node (trackBU) [trackBanner, text=maltCoral] at (2.2, 3.6) {BOTTOM-UP\\ SYNTHESIS};

    \node (cAnchor) [targetNode] at (-2.2, 2.2) {Wikidata Entity:\\ \normalfont\texttt{Q179012}\\ \scriptsize + Rich Metadata};
    \draw [draw=maltCharcoal!30, dashed, line width=1pt] (cAnchor) -- (trackTD);

    \node (cEmerging) [coralNode] at (2.2, 2.2) {Synthesized Entity:\\ Web Project Mgmt\\ \scriptsize + Derived Metadata};
    \draw [draw=maltCoral!40, dashed, line width=1pt] (cEmerging) -- (trackBU);

    \node (hDot1) [skillDot] at (-2.2, 0.6) {};
    \node (hDot2) [skillDot] at (0.0, 0.6) {};
    \node (hDot3) [outlierDot] at (2.2, 0.6) {};
    
    \node (lblh1) [labelNode, below=4pt of hDot1] {Project\\ Management};
    \node (lblh2) [labelNode, below=4pt of hDot2, text width=1.6cm] {Web PM};
    \node (lblh3) [coralLabelNode, below=4pt of hDot3] {Web Project\\ Management};

    \draw [edgeLine] (hDot1) -- (cAnchor.south);
    \node [text=maltCharcoal!70, font=\fontseries{b}\selectfont\scriptsize, anchor=east] at (-2.4, 1.0) {Grounding};

    \draw [edgeLine] (hDot2) -- (cEmerging.south);
    \draw [loopLine] (hDot3) -- (cEmerging.south);
    \node [text=maltCoral, font=\fontseries{b}\selectfont\scriptsize, anchor=west, text width=1.4cm, align=left] at (2.4, 1.4) {Agentic\\ Reflection};

    \node [font=\normalsize\bfseries, text=maltCharcoal] at (0, -0.8) {(C) Proposed Hybrid};
\end{scope}

\end{tikzpicture}
    }
\caption{ \footnotesize Comparison of structural modeling paradigms via a project management narrative. (A) Static Top-Down: Misses emerging specializations due to static ontological boundaries. (B) Unconstrained Bottom-Up: Lacks guardrails, causing semantic fragmentation (separating ``Project Mgmt'' and ``Web PM'') and link hallucinations (``Excel''). (C) Proposed Hybrid: Achieves complete convergence. The baseline concept is anchored to a Wikidata entity node, while agentic reflection synthesizes a distinct sub-entity linked with rich relational metadata that cleanly absorbs related shortcuts (``Web PM'').}\label{fig:paradigm-comparison}
\end{figure}
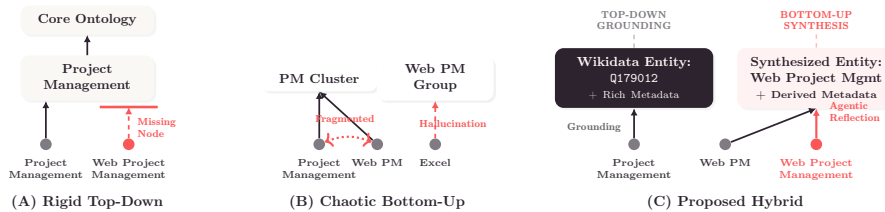

As shown in Figure \ref{fig:paradigm-comparison}, traditional generation methods generally fall into two complementary yet individually limited paradigms. Rigid top-down methods rely on predefined ontologies such as "European Skills, Competences, Qualifications and Occupations" (ESCO) \cite{esco_citation}, which provide structural consistency and high precision but often struggle to capture emerging or highly localized concepts. Conversely, purely generative bottom-up approaches are flexible enough to identify novel trends, but they frequently lack structural constraints. As a result, semantically equivalent concepts may fragment into redundant clusters, while noisy relationships can emerge between otherwise unrelated concepts.

To bridge this gap, we introduce a fully automated, agentic, hybrid pipeline. Our architecture grounds an LLM in an interchangeable Knowledge Graph, instantiated here with Wikidata \cite{vrandecic2014wikidata}, to ensure that the generated entities and their associated metadata remain verifiable and auditable. At the same time, we leverage an iterative multi-agent loop to capture, validate, and structure the emerging "long-tail" skills that formal ontologies have not yet registered. Ultimately, this approach yields a rich knowledge graph from which a streamlined skills taxonomy is derived as a structured child entity.
Specifically, this paper makes the following contributions:

\vspace{0.5mm}
\noindent\textbf{Hybrid Skill Extraction.} We combine structured knowledge graph integration with dynamic, bottom-up parsing to capture both established global standards and emerging long-tail expertise, enriching them with contextual metadata.

\vspace{0.5mm}
\noindent\textbf{Multilingual Standardization.} We enforce structural schema constraints on the underlying model to support consistent multilingual representations across five target languages and reduce hallucinated outputs.

\vspace{0.5mm}
\noindent\textbf{Iterative Self-Refinement.} We introduce a reflection-based multi-agent loop in which agents iteratively analyze and correct intermediate outputs. This enables progressive consolidation of fragmented concepts and improves overall taxonomic coherence.\\

To contextualize these contributions, the remainder of this paper reviews the evolution of automated taxonomy generation, details our pipeline's architecture, and evaluates its real-world performance on our platform data.

\section{Related Work}

Taxonomy creation has shifted from costly expert ontologies and static knowledge bases \cite{gruber1993translation} to automated machine learning methods. Traditional skill extraction models \cite{decorte2021jobbert,skillspan_citation}, and even some recent efficient encoders \cite{context_citation}, primarily produced flat lists susceptible to semantic ambiguity.To establish the relational hierarchies necessary for a true knowledge graph, top-down approaches anchor semantics using Knowledge Graphs, utilizing domain seeds \cite{esco_citation}, automated tree expansion \cite{shen2019hiexpan_1910_08194,zhang2018taxogen,lee2022taxocom}, and LLM-driven ranking and iterative prompting \cite{marchenko2024taxorankconstruct,zeng2024chain}. However, these static methods struggle to adapt to the dynamic vocabularies of modern job markets.

Conversely, bottom-up clustering scales effectively \cite{aggarwal2012survey,clustering_citation} but might lead to uninterpretable labels and missing relational metadata \cite{chang2009reading}. LLM based models overcome this by autonomously structuring categories via abstractive prompting \cite{pham2023topicgpt,wang2023goal}, localized induction \cite{gunn2024creating_2402_12557,gao2025science_2504_13834}, and specialized schemas \cite{sas2024automatic}. While end-to-end methods automate generation \cite{tnt_llm_citation}, their batch processing might lead to fragmented hierarchies rather than cohesive graphs. Recent multi-agent frameworks resolve this through reflection \cite{climb_citation,reflection_citation} and dynamic alignment \cite{kargupta2025taxoadapt_2506_10737}, leveraging complex reasoning heuristics like self-correction \cite{madaan2023self}, prompt optimization \cite{pryzant2023automatic}, and tree search \cite{yao2023tree}. Additionally, LLMs can distill domain knowledge into lightweight downstream classifiers \cite{lee2023making}.

Despite these advances, unconstrained LLMs remain susceptible to hallucinated skills, formatting instability, and bias. We address this via a hybrid architecture: while an LLM drives the pipeline, its reasoning is strictly anchored to a deterministic KG for recognized entities, reserving unconstrained generative reflection solely for unmapped skills and their corresponding relational metadata.

\section{Proposed Approach}

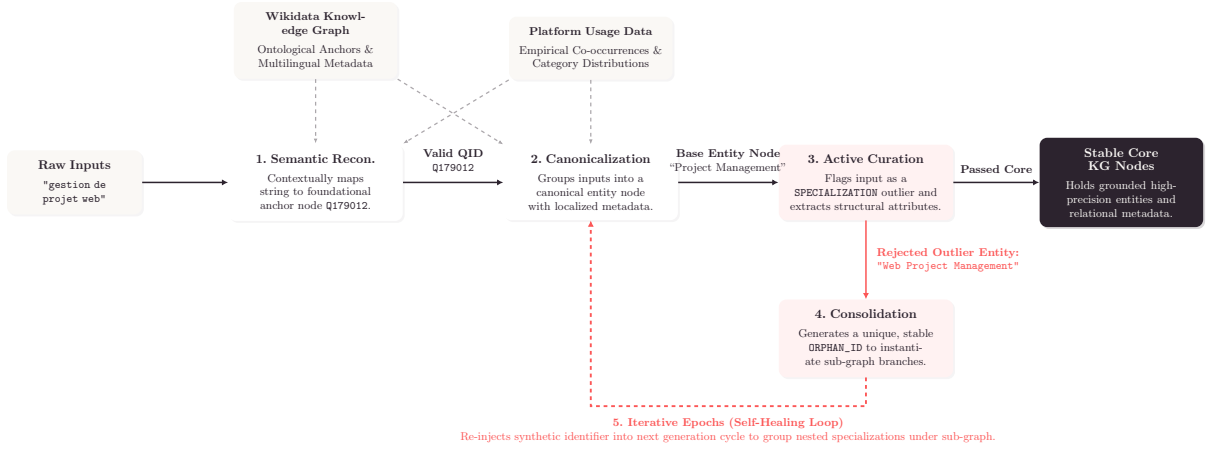
\begin{figure}[htbp]
    \centering
    \resizebox{\linewidth}{!}{%
        \begin{tikzpicture}[
    node distance=1.6cm and 3.0cm,
    baseNode/.style={rectangle, rounded corners=6pt, draw=none, inner sep=8pt, align=center, drop shadow={shadow xshift=1.5pt, shadow yshift=-1.5pt, fill=maltCharcoal!12, opacity=0.4}},
    pipelineNode/.style={baseNode, fill=white, text width=4.3cm, font=\small, text=maltCharcoal},
    contextNode/.style={baseNode, fill=maltCream, text width=4.0cm, font=\small, text=maltCharcoal},
    outputNode/.style={baseNode, fill=maltCharcoal, text width=4.0cm, font=\small\bfseries, text=maltCream, drop shadow={shadow xshift=1.5pt, shadow yshift=-1.5pt, fill=maltCharcoal!30, opacity=0.5}},
    arrowLine/.style={-{Latex[length=5pt, width=5pt]}, draw=maltCharcoal, line width=1.2pt},
    loopLine/.style={-{Latex[length=5pt, width=5pt]}, draw=maltCoral, line width=1.5pt, dashed},
    contextArrow/.style={-{Latex[length=4pt]}, draw=maltCharcoal!40, line width=1pt, dashed}
]

\definecolor{maltCoral}{HTML}{FC5757}       
\definecolor{maltCoralLight}{HTML}{FFF2F2}  
\definecolor{maltCharcoal}{HTML}{2C232E}   
\definecolor{maltCream}{HTML}{FAF8F5}

\node (input) [pipelineNode, fill=maltCream, text width=3.2cm] at (0,0) {
    {\normalsize\bfseries Raw Inputs}\\
    \vspace{4pt}
    \texttt{"gestion de projet web"}
};

\node (step1) [pipelineNode, right=2.4cm of input] {
    {\normalsize\bfseries 1. Semantic Recon.}\\
    \vspace{4pt}
    Contextually maps string to foundational anchor node \texttt{Q179012}.
};

\node (step2) [pipelineNode, right=2.8cm of step1] {
    {\normalsize\bfseries 2. Canonicalization}\\
    \vspace{4pt}
    Groups inputs into a canonical entity node with localized metadata.
};

\node (step3) [pipelineNode, fill=maltCoralLight, right=2.8cm of step2] {
    {\normalsize\bfseries 3. Active Curation}\\
    \vspace{4pt}
    Flags input as a \texttt{SPECIALIZATION} outlier and extracts structural attributes.
};

\node (output) [outputNode, right=2.4cm of step3] {
    {\normalsize\bfseries Stable Core KG Nodes}\\
    \vspace{4pt}
    \normalfont Holds grounded high-precision entities and relational metadata.
};

\node (wikidata) [contextNode, above=1.8cm of step1] {
    \textbf{Wikidata Knowledge Graph}\\
    \vspace{4pt}
    Ontological Anchors \&\\ Multilingual Metadata
};

\node (maltdata) [contextNode, above=1.8cm of step2] {
    \textbf{Platform Usage Data}\\
    \vspace{4pt}
    Empirical Co-occurrences \&\\ Category Distributions
};

\node (step4) [pipelineNode, fill=maltCoralLight, below=2.2cm of step3] {
    {\normalsize\bfseries 4. Consolidation}\\
    \vspace{4pt}
    Generates a unique, stable \texttt{ORPHAN\_ID} to instantiate sub-graph branches.
};

\draw [arrowLine] (input.east) -- (step1.west);

\draw [arrowLine] (step1.east) -- node[above=4pt, font=\small, align=center, text=maltCharcoal] {
    \textbf{Valid QID}\\ \texttt{Q179012}
} (step2.west);

\draw [arrowLine] (step2.east) -- node[above=4pt, font=\small, align=center, text=maltCharcoal] {
    \textbf{Base Entity Node}\\ ``Project Management''
} (step3.west);

\draw [arrowLine] (step3.east) -- node[above=4pt, font=\small\bfseries, text=maltCharcoal] {Passed Core} (output.west);

\draw [arrowLine, draw=maltCoral] (step3.south) -- node[right=5pt, font=\small\bfseries, text=maltCoral, align=left] {
    Rejected Outlier Entity:\\ \texttt{"Web Project Management"}
} (step4.north);

\draw [loopLine] (step4.south) -- ++(0,-0.8) coordinate (LoopCorner1) -- 
    node[below=5pt, font=\small\bfseries, text=maltCoral, align=center] {
        5. Iterative Epochs (Self-Healing Loop)\\
        \normalfont\small Re-injects synthetic identifier into next generation cycle to group nested specializations under sub-graph.
    } 
    (LoopCorner1 -| step2.south) -- (step2.south);

\draw [contextArrow] (wikidata.south) -- (step1.north);
\draw [contextArrow] (wikidata.south east) -- (step2.north west);
\draw [contextArrow] (maltdata.south west) -- (step1.north east);
\draw [contextArrow] (maltdata.south) -- (step2.north);

\end{tikzpicture}
    }
\caption{\footnotesize The 5-stage hybrid pipeline architecture, traced via the project management example. The system grounds the model in Wikidata (Stage 1) and uses agentic reflection (Stage 3) to capture outliers. These are converted into stable synthetic identifiers to instantiate new sub-graph entities (Stage 4) and rerouted (Stage 5) for continuous, autonomous self-healing of the knowledge graph structure.}
    \label{fig:full-pipeline}
\end{figure}

We designed a hybrid, multi-agent pipeline to overcome the limitations of rigid top-down ontologies and ungrounded bottom-up generation. Unlike top-down methods that map unstructured inputs to a static seed ontology, our system dynamically constructs its own custom, evolving knowledge graph derived directly from empirical data. Conversely, unlike pure bottom-up methods that rely on unconstrained generative clustering, we enforce strict semantic grounding by anchoring these emerging clusters to factual Knowledge Graph entities and capturing their relational metadata. As illustrated in Figure \ref{fig:full-pipeline}, the system operates as an iterative loop rather than a strictly linear execution pipeline, processing batches through five distinct stages. By utilizing Wikidata as a semantic anchor, chosen for its broad multilingual domain coverage enabled by its massive open-source scale, and constraining Gemini 1.5 Flash with strict schemas, we prevent hallucinations while preserving linguistic variants and contextual structure. We specifically selected this lightweight model for its cost-efficiency and immediate availability within our infrastructure. Because our pipeline relies on strict external grounding rather than complex internal reasoning, these results could likely be replicated using comparable open-source small language models. Crucially, this architecture is not tied to a single semantic structure and could easily integrate alternative or combined external databases. The following subsections detail the mechanics of each phase, demonstrating how the pipeline enables continuous, autonomous updates as novel expertises and their structural relationships emerge in the market.

\subsection{Reconciliation: Grounding Raw Skills in Wikidata}

\noindent\colorbox{black!5}{%
    \parbox{\dimexpr\linewidth-2\fboxsep\relax}{%
        \tiny \textbf{Trace Example --- Stage 1 (Reconciliation):} 
        \textbf{In:} Raw Text (\texttt{"gestion de projet web"}) + Malt  Context.  
        \textbf{Out:} Linked Anchor (\texttt{Q179012}) + Validation Flags (\texttt{is\_skill: true}, \texttt{is\_compound: false})
    }%
}
\vspace{0.5mm}

As the first step in this pipeline, the Reconciliation phase maps noisy, unstandardized, and multilingual skill mentions to stable, unambiguous Wikidata entity identifiers (QIDs), which serve as the foundational anchor nodes for our knowledge graph. This process begins by extracting the top ten Wikidata entries linked to the raw input text, alongside metadata such as labels, descriptions, and regional variants across the five target languages. Because raw user mentions are frequently ambiguous or completely devoid of context when analyzed in isolation, this baseline retrieval must be structurally enriched with domain-specific semantic neighborhoods and high-level category distributions.

At Malt, this enrichment queries the platform's user profile graph. Upon skill ingestion, the engine extracts and appends two empirical features from the freelancer's history: the top fifteen (an empirical threshold selected to maximize semantic signal while filtering out long-tail profile noise) co-occurring peer skills and the top five overarching professional categories.

From this enriched context, the LLM executes a disambiguation step to select the best corresponding QID. To ensure structured outputs, the LLM must output a JSON payload containing the selected QIDs, confidence scores, step-by-step reasoning, and critical boolean flags indicating whether the input is a valid professional skill (\texttt{is\_skill}) and if it contains multiple skills (\texttt{is\_compound}). 

This dual validation approach, combining Wikidata's semantic recall with the LLM's context-aware precision, reduces potential hallucinated mappings. Ultimately, by evaluating candidate labels in all target languages simultaneously, the system ensures that non-English skills are reliably anchored to the exact same global QID as their English counterparts.  

\subsection{Canonicalization: Clustering and Preferred Label Generation}

\noindent\colorbox{black!5}{%
    \parbox{\dimexpr\linewidth-2\fboxsep\relax}{%
        \tiny \textbf{Trace Example --- Stage 2 (Canonicalization):} 
        \textbf{In:} Resolved Entity Cluster (\texttt{Q179012}) + Historical Usage Logs.
        \textbf{Out:} Multilingual Label Mapping (\texttt{EN:} "Project Management" \texttt{[wikidata]}, \texttt{FR:} "Gestion de Projet" \texttt{[malt]})
    }%
}
\vspace{0.5mm}

Following reconciliation, the second phase groups validated inputs by their unique QID combinations to generate human-readable, multilingual canonical names. Provided with context such as Wikidata descriptions and empirical usage distributions within the freelance platform, the LLM synthesizes localized preferred labels across five target languages. To mitigate clustering noise, the prompt enforces a strict fallback hierarchy: the model must prioritize empirical platform usage, fall back to official ontological titles, and synthesize novel labels only when necessary. To maintain algorithmic explicability, each label is explicitly tagged with its resulting provenance (\texttt{malt}, \texttt{wikidata}, or \texttt{generative}).

\subsection{Curation: Semantic Validation and Reflexion}

\noindent\colorbox{black!5}{%
    \parbox{\dimexpr\linewidth-2\fboxsep\relax}{%
        \tiny \textbf{Trace Example --- Stage 3 (Curation):} 
        \textbf{In:} Combined Cluster Member (\texttt{"gestion de projet web"}) vs. Core Candidate ("Project Management"). 
        \textbf{Out:} Status: \texttt{REJECTED} (Criterion: \texttt{SPECIALIZATION}) $\rightarrow$ Output Target: \texttt{suggested\_pref\_label: "Web Project Management"}
    }%
}

\vspace{0.5mm}

To prevent disjointed or overly granular skills within canonicalized nodes, an LLM-powered Curation agent validates the strict equivalence of every raw skill against its broader entity grouping and selected preferred label. If a skill is non-equivalent, the model outputs a structured output detailing one of seven granular rejection criteria (\texttt{AMBIGUOUS}, \texttt{SPECIALIZATION}, \texttt{SEMANTIC\_MISMATCH}, \texttt{NOT\_A\_SKILL}, \texttt{METHODOLOGY}, \texttt{CONTEXT}, \texttt{SUB\_TASK}). For pipeline tracking and evaluation purposes, these semantic rejections (aside from \texttt{NOT\_A\_SKILL}) are aggregated under the broader \texttt{NODE\_CURATION} status. Crucially, a \texttt{suggested\_pref\_label} is generated to define what the rejected concept should actually be named. This contextual rejection reason is retained as metadata for subsequent consolidation tasks.

Following outlier extraction, the agent re-evaluates and refines the core node's canonical label using strictly the accepted subset. This bifurcates the data into tightly curated baseline entities and a structured "Orphan" queue for iteration, which will be processed later as explained in section \ref{sec:iteration}. As a safeguard, any node with a rejection rate exceeding 50\% is automatically flagged for human-in-the-loop review to prevent cascading systemic errors.

\subsection{Consolidation: Cross-Batch Deduplication}

\noindent\colorbox{black!5}{%
    \parbox{\dimexpr\linewidth-2\fboxsep\relax}{%
        \tiny \textbf{Trace Example --- Stage 4 (Consolidation):} 
        \textbf{In:} Cross-Batch Incoming Candidate (\texttt{"Web PM"}) vs. Active Orphan Taxonomy Target (\texttt{"Web Project Management"}).
        \textbf{Out:} Decision: \texttt{MERGE} $\rightarrow$ Surviving Entity Destination ID: \texttt{hash("Web Project Management")}
    }%
}

\vspace{0.5mm}

To maintain global structural consistency across incremental runs, the Consolidation phase merges overlapping sub-graphs. To avoid the computational explosion of exhaustive pairwise comparisons, we employ an asymmetric bootstrapping strategy. The system initializes from a blank state, where the first processed batch establishes the foundational knowledge graph. In all subsequent runs, newly generated nodes are strictly compared against this continuously growing, established baseline. Before LLM evaluation, lightweight heuristics flag potential merges between incoming and established entities based on \textbf{member intersection} (shared raw skills) or \textbf{lexical similarity} (low edit distance between preferred labels). Once flagged, the LLM evaluates the combined metadata of these pairs to output a structural decision (\texttt{MERGE} or \texttt{KEEP\_SEPARATE}), designating a surviving ID if merged. Logging these decisions creates a cache that prevents redundant re-evaluations in future Epochs.

\subsection{Iteration: Orphan Recovery and Convergence}
\label{sec:iteration}

\noindent\colorbox{black!5}{%
    \parbox{\dimexpr\linewidth-2\fboxsep\relax}{%
        \tiny \textbf{Trace Example --- Stage 5 (Iteration):} 
        \textbf{In:} Verified Isolated Structural Orphan Queue + Validated Label Target .
        \textbf{Out:} Sub-graph Generation: Instantiates stable sub-branch linked to \texttt{Q179012} via deterministic cryptographic label routing.
    }%
}

\begin{figure}[htbp]
    \centering
    \resizebox{0.75\linewidth}{!}{%
        \newcommand{\codeText}[1]{{\ttfamily\small\textcolor{maltCharcoal}{#1}}}

\begin{tikzpicture}[
    node distance=1.2cm and 2.6cm, 
    baseNode/.style={rectangle, rounded corners=6pt, inner sep=8pt, align=center, font=\small, draw=none, text=maltCharcoal, drop shadow={shadow xshift=1.5pt, shadow yshift=-1.5pt, fill=maltCharcoal!12, opacity=0.4}},
    startNode/.style={baseNode, fill=maltCream, text width=4.2cm},
    decisionNode/.style={diamond, aspect=2.2, draw=none, fill=maltCream, align=center, font=\small\bfseries, inner sep=6pt, text=maltCharcoal, drop shadow={shadow xshift=1.5pt, shadow yshift=-1.5pt, fill=maltCharcoal!12, opacity=0.4}},
    successNode/.style={baseNode, fill=maltCharcoal, text width=3.8cm, text=maltCream},
    reflexionNode/.style={baseNode, fill=white, text width=5.0cm},
    arrowLine/.style={-{Latex[length=5pt, width=5pt]}, draw=maltCharcoal, line width=1.2pt},
    loopLine/.style={-{Latex[length=5pt, width=5pt]}, draw=maltCoral, line width=1.5pt, dashed},
    codeText/.style={font=\ttfamily\small, text=maltCharcoal}
]

\definecolor{maltCoral}{HTML}{FC5757}
\definecolor{maltCoralLight}{HTML}{FFF2F2}
\definecolor{maltCharcoal}{HTML}{2C232E}
\definecolor{maltCream}{HTML}{FAF8F5}

\node (step1) [startNode] {
    {\normalsize\bfseries Step 1: Canonical Entity Node}\\
    \vspace{4pt}
    Input Node \texttt{Q192253} (``Project Management'')
};

\node (step2) [decisionNode, below=1.4cm of step1] {
    Step 2:\\ LLM Validation\\
    \normalfont\small Is skill equivalent?
};

\node (step3yes) [successNode, left=2.2cm of step2] {
    {\normalsize\bfseries Step 3 (Yes): Core KG Node}\\
    \vspace{4pt}
    \normalfont Accepts stable generic node:\\ ``Project Management''
};

\node (step3no) [reflexionNode, right=2.2cm of step2, yshift=-0.5cm] {
    {\normalsize\bfseries Step 3 (No): Rejection Analysis}\\
    \vspace{4pt}
    Flags ``gestion de projet web'' as a \texttt{SPECIALIZATION}
};

\node (step4) [reflexionNode, below=0.8cm of step3no] {
    {\normalsize\bfseries Step 4: Label Generation}\\
    \vspace{4pt}
    Synthesizes \codeText{suggested\_pref\_label}:\\ ``Web Project Management''
};

\node (step5) [reflexionNode, below=0.8cm of step4] {
    {\normalsize\bfseries Step 5: Orphan Node Creation}\\
    \vspace{4pt}
    Uses suggested label to define a unique, stable \codeText{ORPHAN\_ID}
};

\coordinate (boxTopLeft) at ($(step3no.north west) + (-0.4, 0.6)$);
\coordinate (boxBottomRight) at ($(step5.south east) + (0.4, -0.4)$);

\draw [draw=none, rounded corners=8pt, fill=maltCoralLight, drop shadow={shadow xshift=1.5pt, shadow yshift=-1.5pt, fill=maltCharcoal!10, opacity=0.3}] 
    (boxTopLeft) rectangle (boxBottomRight);
        
\node [font=\small\bfseries, text=maltCoral, anchor=north west] at ($(boxTopLeft) + (0.15, -0.15)$) {Agentic Reflection Phase};

\node [reflexionNode] at (step3no) {
    {\normalsize\bfseries Step 3 (No): Rejection Analysis}\\
    \vspace{4pt}
    Flags ``gestion de projet web'' as a \texttt{SPECIALIZATION}
};
\node [reflexionNode] at (step4) {
    {\normalsize\bfseries Step 4: Label Generation}\\
    \vspace{4pt}
    Synthesizes \codeText{suggested\_pref\_label}:\\ ``Web Project Management''
};
\node [reflexionNode] at (step5) {
    {\normalsize\bfseries Step 5: Orphan Node Creation}\\
    \vspace{4pt}
    Uses suggested label to define a unique, stable \codeText{ORPHAN\_ID}
};

\draw [arrowLine] (step1) -- (step2);

\draw [arrowLine] (step2) -- node[above=3pt, font=\small\bfseries, text=maltCharcoal] {Yes} (step3yes);

\draw [arrowLine, draw=maltCoral] (step2.east) -- ++(1.0,0) |- node[above=3pt, pos=0.2, font=\small\bfseries, text=maltCoral] {No (Reject)} (step3no.west);

\draw [arrowLine, draw=maltCharcoal!40] (step3no) -- (step4);
\draw [arrowLine, draw=maltCharcoal!40] (step4) -- (step5);

\draw [loopLine] (step5.east) -- ++(1.0,0) |- 
    node[pos=0.25, right=6pt, font=\small\bfseries, text=maltCoral, align=left] {
        Step 6: Route to Epoch $N+1$\\
        \normalfont\small Forced grouping of nested\\
        \normalfont\small specialization into sub-graph
    } 
    (step1.east);

\end{tikzpicture}
    }
\caption{\footnotesize The Agentic Reflection and Orphan Lifecycle. Rejected skills trigger an active reflection loop (Steps 3--5) where the LLM justifies the outlier status and synthesizes a \texttt{suggested\_pref\_label}. This generates a stable synthetic \texttt{ORPHAN\_ID} routed into the next Epoch for autonomous self-healing of the knowledge graph layout.}
    \label{fig:reflexion-lifecycle}
\end{figure}
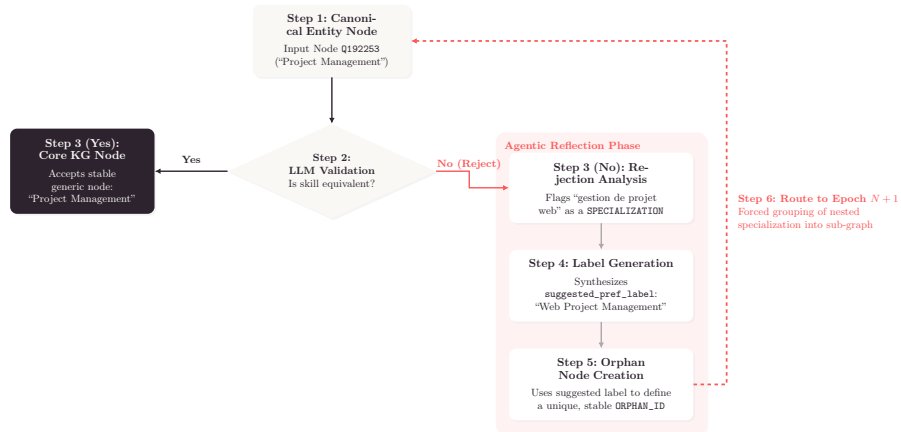

This iterative loop primarily serves to fix granularity gaps inherited from the baseline anchor graph. For example, both "Project Management" and "Web Project Management" might initially map to the same broad Wikidata entity. To preserve graph coherence, the curation phase flags "Web Project Management" as an outlier while keeping the main node broad. Instead of discarding this niche specialization, the pipeline captures it as an "orphan" to generate the fine-grained relational edges and specialized nodes that Wikidata lacks natively.

As illustrated in Figure \ref{fig:reflexion-lifecycle}, these orphans are re-processed using a recursive routing loop. Each orphan is assigned a synthetic identifier derived directly from its LLM-generated \texttt{suggested\_pref\_label}. Because identical concepts receive the exact same suggested label from the model, this mechanism naturally groups separate but matching specializations together into a unified sub-graph in the subsequent Epoch. This loop repeats until the orphan queue is empty, achieving full semantic convergence.

\section{Evaluation}

We evaluated our pipeline using a proprietary dataset of unstructured, multilingual expertise declarations from the Malt freelancing platform. This dataset reflects chaotic, real-world labor market dynamics—featuring a severe long-tail distribution of highly niche, emerging, or misspelled jargon—providing a robust stress test compared to static theoretical ontologies. Because HR matching engines require strict reliability, model performance is evaluated against a curated Wikidata gold standard. As this research represents ongoing work, our current evaluation focuses strictly on the initial retrieval and pre-consolidation phases; the final post-consolidation phase was recently introduced to the pipeline and has yet to be formally benchmarked.

\subsection{Vocabulary Coverage and Semantic Compression}

The normalization pipeline processed an initial vocabulary of 36,037 raw expertise strings. The reconciliation engine successfully resolved 27,743 of these inputs, achieving a Global Coverage rate of 77\% (the percentage of valid inputs successfully mapped to a knowledge graph node). The remaining 8,294 unmapped entries were flagged by the curation layer as non-skills or semantic noise (see Table \ref{tab:rejection_distribution}).The 27,743 mapped variations were grouped into 15,010 semantic groupings, which were further streamlined into 13,298 canonical skill nodes. This represents a compression rate of 52.1\% ($1-[13,298/27,743]$), significantly reducing downstream redundancy. Alongside, the Average Skills per Node ($ASpN$) metric, which stands at 2.08 variations per canonical node,  further demonstrates our vocabulary consolidation efficiency. Importantly, the knowledge graph maintains perfect cross-lingual symmetry: 100\% of the 13,298 canonical concept nodes are fully supported across all five target locales (fr, en, de, nl, es). This generates exactly 66,490 standardized preferred labels ($13,298\times5$), ensuring uniform matching regardless of the user's interface language. Empirical platform data reveals a strong Pareto distribution: the top 1,000 canonical skills account for 82.74\% of platform usage volume, and the top 5,000 capture 97.25\%. Interpreting this requires distinguishing between the lexical long-tail (typographical noise and redundant expression variants) and the semantic long-tail (rare, highly specialized, or emerging capabilities). While our pipeline aggressively filters and compresses the unmanaged lexical long-tail to eliminate marketplace redundancy, it systematically preserves the semantic long-tail through agentic reflection, yielding a highly comprehensive and nuanced final knowledge graph of 13,298 standardized concept nodes.

\subsection{Quantitative Precision against Gold Standard}
Evaluated against a hand-annotated gold standard, independently curated by five domain experts without overlap, the pre-consolidation pipeline achieved a global baseline Alignment Coverage of 79.7\% (the overall proportion of inputs successfully and correctly mapped to the gold standard) and a Found Coverage of 84.9\% (measuring precision strictly on the subset of inputs where the model actually attempted a retrieval). Due to the semantic ambiguity of unmanaged inputs, the global Wrong Guess Rate (WGR) was 19.1\%. Given the chaotic nature of user-generated profile text, this error rate remains highly competitive and is mitigated by downstream curation. As shown in Table \ref{tab:domain_coverage}, performance varies by domain. Highly structured domains like Video Games (91.8\% Found Coverage) outperformed softer, more subjective fields like Communication (81\% Found Coverage), where shifting jargon and conceptual overlap complicate alignment.

\subsection{Provenance and Structural Purity}

To ensure explicability, we track provenance metadata across all 66,490 preferred labels. Accounting for source overlap, 80.65\% of the knowledge graph is strictly anchored in factual, real-world data (comprising 67.28\% empirical platform usage and 22.08\% Wikidata titles). The generative engine synthesized only the remaining 19.35\% to resolve emerging concepts and isolated language gaps, proving the structure is rooted in empirical reality rather than ungrounded model hallucinations.

Finally, we measured structural coherence using the Outlier Rate (skills manually removed during human review vs. total generated sub-graphs). Qualitative audits revealed highly cohesive node groupings, with marginal error rates (0.01 to 0.06 outliers per sub-graph). This purity stems from the Active Curation agent, which aggressively isolates semantic noise before ingestion, successfully blocking 13.7\% of rejected inputs via the NOT\_A\_SKILL filter (Table \ref{tab:rejection_distribution}).

\begin{table}[htbp]
    \centering
    \begin{minipage}[t]{0.52\textwidth}
        \centering
        \caption{Baseline Reconciliation Coverage}
        \label{tab:domain_coverage}
        \scriptsize
        \begin{tabular}{lcc}
        \hline
        \textbf{Domain} & \textbf{Alignment Cov.} & \textbf{Found Cov.} \\
        \hline
        Video Games & 88.1\% & 91.8\% \\
        Industrial Eng. & 86.0\% & 89.9\% \\
        Data \& Analytics & 83.5\% & 88.7\% \\
        Tech / Software & 81.7\% & 86.8\% \\
        Marketing & 76.7\% & 82.4\% \\
        Communication & 75.4\% & 81.0\% \\
        \hline
        \textbf{Global Baseline} & \textbf{79.7\%} & \textbf{84.9\%} \\
        \hline
        \end{tabular}
    \end{minipage}\hfill
    \begin{minipage}[t]{0.44\textwidth}
        \centering
        \caption{Active Curation Rejections {\scriptsize \\ \textit{Note: \texttt{NODE\_CURATION} aggregates Section 3.3 criteria.}}}
        \label{tab:rejection_distribution}
        \scriptsize
        \begin{tabular}{lc}
        \hline
        \textbf{Rejection Reason} & \textbf{\% of Total Rejected} \\
        \hline
        \texttt{SCORE\_REJECTED} & 31.9\% \\
        \texttt{NODE\_CURATION} & 28.1\% \\
        \texttt{NOT\_A\_SKILL} & 13.7\% \\
        \texttt{WIKIDATA\_NOT\_FOUND} & 11.4\% \\
        \texttt{NOT\_IN\_NODE} & 10.7\% \\
        \texttt{ERROR\_NOT\_KNOWN} & 4.2\% \\
        \hline
        \end{tabular}
    \end{minipage}
\end{table}

\section{Discussion and Future Work}

While this pipeline provides a structured representation of the underlying data, it represents an initial step toward modeling the complexity of modern labor markets. Such structured representations are a prerequisite for building robust HR analytics systems. Looking ahead, a key objective is to identify emerging skill  signals in real-time, enabling downstream analysis of labor market dynamics and supporting workforce planning applications. A streamlined skill taxonomy, extracted directly from this underlying knowledge graph, is already integrated into the platform's candidate-matching algorithms. A primary benefit of this deployment is that the graph provides an intermediate abstraction layer that significantly improves the interpretability and auditability of our automated matching systems. While relying on Wikidata as a primary anchor introduces limitations, such as a lag in capturing niche HR jargon and structural inconsistencies due to its generalist nature, our hybrid approach mitigates this. The bottom-up orphan recovery loop acts as a safety net, autonomously structuring the emerging long-tail skills that Wikidata natively misses. As the system scales, it is important to explicitly address potential representational biases. Large language and embedding-based models often exhibit English-centric tendencies, which can lead to over-normalization of non-English occupational structures and a reduced fidelity of locale-specific distinctions. 
Our goal is to mitigate these effects by preserving meaningful cross-lingual variation in occupational and skill representations.
Addressing gender-related bias is also critical. In highly inflected languages such as French and German, occupational and skill terms are often gender-marked. We aim for the underlying knowledge graph to support the seamless mapping of these gendered variants to shared underlying occupational concepts, while preserving their distinct linguistic forms for accurate representation and equitable matching.
Finally, to evaluate the robustness of the system  at scale, our evaluation roadmap will focus on three key areas:

\vspace{0.5mm}
\noindent\textbf{Extended Evaluations.} While parts of our pipeline, particularly the consolidation phase, are already implemented, they require further evaluation. We will benchmark the consolidated graph structure against existing standards (like ESCO) and other AI approaches (like TnT-LLM \cite{tnt_llm_citation} or CLIMB \cite{climb_citation}), while assessing the system’s ability to incorporate new skills and the effectiveness of the human-in-the-loop components.
    
\vspace{0.5mm}
\noindent\textbf{Cost and Scalability.} Large-scale deployment of LLM-based pipelines introduces significant computational costs. We will measure token consumption, processing latency, and optimization strategies required to ensure operational scalability.
    
\vspace{0.5mm}
\noindent\textbf{Failure Analysis.} To support responsible deployment, we will analyze failure cases by tracking manual intervention rates, categorizing systematic mapping errors, and evaluating performance degradation under high-volume concept comparison scenarios.

\section{Conclusion}
This ongoing work introduces a hybrid, multi-agent architecture designed to build skills knowledge graphs that are both rigorously structured and highly adaptable. By grounding a large language model in Wikidata, our pipeline effectively parses multilingual free-text while keeping the underlying entity data reliable and easy to audit. Through continuous curation and automated reflection, the system successfully captures the niche and emerging long-tail instances that traditional, static models often miss. Ultimately, this approach creates a living knowledge graph capable of keeping pace with the rapid technological changes and linguistic shifts of the modern freelance market. While the overarching framework operates as a rich, metadata-driven knowledge graph, its hierarchical output can be easily downstreamed as a clean, structured skills taxonomy.  
\bibliographystyle{splncs03}
\bibliography{biblio}

\appendix

\section*{Supplementary Material}

To ease reproducibility while respecting the strict confidentiality of our proprietary platform data and internal infrastructure, we provide in this appendix the templates of all the prompts for all five stages of the pipeline. These prompts were executed with Gemini 1.5 Flash with temperature set to 0 and structured output defined. Because our architecture is intrinsically dataset-agnostic and anchored to public Wikidata entities, researchers can readily implement and benchmark this exact pipeline using open-source labor market datasets.

\section{Reconciliation details}
This section details the initial phase of the pipeline, where noisy, unstandardized skill mentions are mapped to stable Wikidata entity identifiers (QIDs). As demonstrated in the prompt below, the model is strictly grounded using enriched platform context—specifically, empirical co-occurring peer skills and overarching professional categories—to accurately disambiguate and anchor the input while mitigating hallucinations.

\begin{tcolorbox}[academicexample, breakable, enhanced]
\begin{lstlisting}
You are a highly precise Skill Entity Linking agent. Your mission is to analyze a freelancer's stated expertise, determine if it qualifies as a professional skill, and link it to the most relevant Wikidata item(s) from a provided list of candidates.

### Primary Directives:
1. **Classify the Expertise**: First, determine if the input word "{expertise}" is a skill and whether it is a single or compound skill.
2. **Link the Skill**: If it is a skill, select the best matching Wikidata QID(s) from the candidates using the strict Selection Rules below.

### Definition of a Skill:
- **What IS a skill**: Specific, learnable professional abilities. Includes technical tools, frameworks, programming languages, methodologies, and specific domains.
- **What is NOT a skill**: 
  - General personal attributes or soft skills (e.g., "Hard worker", "Motivated").
  - Levels of seniority or units of time (e.g., "10 years experience", "Senior").

### Candidate Selection Rules (CRUCIAL):
Analyze the provided Wikidata candidates carefully. You must navigate the following edge cases:
1. **Direct Match Principle**: Keep ONLY QIDs that represent the expertise directly, or represent a legitimate component part of a compound expertise.
2. **The Overlap Rule (Deduplication)**: If multiple QIDs refer to the exact same concept or the same part of the expertise, **keep only the single best-matching one** and discard the rest. Do not return 3 different QIDs that all mean the same thing.
3. **The Compound / Combination Rule**: 
   - Freelancers often type compound skills (e.g., "React.js & Node.js").
   - Sometimes a skill is a combination of concepts and Wikidata is too fine-grained.
   - If a single QID does not cover the full signal of the expertise, you MUST select a combination of QIDs that preserves the full signal. Prefer a combination over a single partial match.
4. **The Directionality Rule**: 
   - If the compound skill represents a directional process where the order of items strictly matters (e.g., "English to French translation", "Figma to React", "Data migration from Oracle to Postgres"), you MUST set "is_directional": true.
   - For standard combinations where order doesn't matter (e.g., "React and Node.js"), set it to false.
5. **MULTILINGUAL UNIFICATION RULE (Anti-Splitting)**: 
   - Wikidata QIDs are language-agnostic concepts. Your goal is to map exact translations to the SAME primary universal QID.
   - If the input expertise is in a non-English language (e.g. "Réseaux sociaux" in French), map it to the primary global QID for that concept (which is usually anchored by the English standard, e.g. "Social Media"). 
   - DO NOT select a secondary, narrower QID just because its translated label is a closer literal match. Force direct translations to converge on the same central QID to avoid language siloing.
6. **CRITICAL LANGUAGE RULE (False Friends)**: 
   - Beware of "False Friends" (Faux amis) across languages. Do not map a foreign word to an English Wikidata concept just because they are spelled similarly if the professional meaning is different.
   - Example: The French "Rédaction" means "Copywriting/Writing", it does NOT mean the English "Redaction/Censoring". Prioritize the semantic meaning used in a freelance marketplace context.

### SPECIFICITY RULE: 
If a freelance skill mentions both a broad category and a specific framework (e.g., 'Méthode Agile Scrum'), do NOT return multiple QIDs. You MUST return ONLY the QID of the most specific, primary concept (e.g., only the QID for 'Scrum'). Never create composite QID lists unless the skill is truly a directional mapping (like translating from one language to another).

### Step-by-Step Instructions:
1. **Analyze "{expertise}"**: Based on the definitions above, decide if it's a skill.
2. **Identify Compound Skills**: If it IS a skill, determine if it is a compound skill.
3. **Evaluate Candidates**: Carefully evaluate each candidate QID using the Selection Rules above. Use all provided context (co-occurring skills, job categories, Wikidata info).
4. **Select & Score**:
   - Choose the best QID(s). For a compound skill, you MUST return one entry for each distinct skill identified.
   - For each selected QID, provide a score and a concise reasoning.

### Scoring Rubric:
- **0.0**: The best matching wikidata item(s) is not really a match. It is totally irrelevant for the provided skill.
- **0.4**: Moderate match. Maybe the perfect wikidata item doesn't exist or was not provided.
- **0.7**: Strong match. The wikidata item captures the concept well.
- **1.0**: Perfect match. No other wikidata item or concept will be better.

### Reasoning Instructions:
- Your reasoning must be concise (1-2 sentences).
- **Justify your decision by referencing the provided context**.

### CONTEXT FOR THE TASK
### Co-occurring skills for "{expertise}":
{co_occurrences}

### Job categories for "{expertise}":
{categories}

### Wikidata item candidates for "{expertise}":
{candidates}
\end{lstlisting}
\end{tcolorbox}
\vspace{-2mm}
\captionof{figure}{\footnotesize Semantic Reconciliation Prompt}
\label{fig:reconciliation-prompt}
\vspace{4mm}

\section{Canonicalization details}
Following reconciliation, the canonicalization phase groups validated inputs by their resolved QIDs to synthesize localized, human-readable preferred labels across the five target languages. To handle the diverse complexity of the data, the system utilizes three distinct prompts depending on the entity type: single Wikidata concepts, compound concepts requiring directional logic, and completely novel orphaned skills that require synthetic identifiers.

\begin{tcolorbox}[academicexample, breakable, enhanced]
\begin{lstlisting}
You are an expert **Multilingual Taxonomy Linguist and HR Specialist**.
Your ONLY task is to analyze a single Wikidata concept and its associated raw skills to determine the best Preferred Label (`pref_label`) for each target language.

---

### PRIMARY DIRECTIVES

1. **Preferred Label Selection**
   - Prioritize the most **frequently used** label from the Malt usage counts if it is professional.
   - Fallback to the **Wikidata label** if the usage-based terms are ambiguous, informal, or inappropriate.
   - Create a **Synthetic** (new) professional label if neither accurately describes the core concept.

2. **Provenance Tracking (`source`)**
   - For every localized label, specify its origin:
     - `"malt"`: based on an existing high-usage profile expertise.
     - `"wikidata"`: selected the official Wikidata description or language variant.
     - `"gemini"`: you synthesized a brand new term yourself.

---

### INPUT DATA

**Target Languages:** {", ".join(languages)}

**1. Associated Wikidata Concept:**
{wd_text}

**2. Skills in this Group (Use for context to understand the community usage):**
{malt_text_joined}
\end{lstlisting}
\end{tcolorbox}
\vspace{-2mm}
\captionof{figure}{\footnotesize Canonicalization Prompt - Single QID}
\label{fig:canonicalization-single-prompt}
\vspace{4mm}

\begin{tcolorbox}[academicexample, breakable, enhanced]
\begin{lstlisting}
You are an expert **Multilingual Taxonomy Linguist and HR Specialist**.
Your ONLY task is to analyze a combination of Wikidata concepts and the raw skills associated with them to determine the best unified Preferred Label (`pref_label`) and final Concept Identifier.

---

### PRIMARY DIRECTIVES

1. **Compound Label Selection**
   - You MUST choose a Preferred Label that is coherent with the **majority of the high-usage skills** within the cluster.
   - **Directionality Rule (CRITICAL):** If `Is Directional Process` is `True` in the Input Data, the order of the concepts strictly matters (e.g., a migration or translation from Concept A to Concept B). Your label MUST reflect this directional relationship (e.g., "English to French Translation" instead of "English and French").
   - If `Is Directional Process` is `False`, treat it as a standard combination and look for a standard industry umbrella term (e.g., "MERN Stack").

2. **Define Logic for the Identifier**
   - The Baseline Composite ID groups the provided QIDs (e.g., `{qid_combo}`). 
   - If you modify or add to this identifier, use `&` (AND) for integrated practices, and `|` (OR) for independent traits.

3. **Synthetic Identifiers**
   - You must exclusively use the provided Wikidata QIDs whenever possible. 
   - However, if the majority of the skills introduce a critical, distinct concept that is NOT covered by any provided QID, you MUST invent a concise, uppercase English textual identifier and append it (e.g., `{qid_combo} & WEB` or `{qid_combo} | IT`). Use the exact same textual identifier across all languages for consistency.

4. **Create Preferred Labels**
   - Localize names for the final composite identifier (including those with synthetic textual IDs appended). Prioritize usage counts, fallback to Wikidata, or synthesize a professional HR term if needed.

5. **Provenance Tracking (`source`)**
   - For every localized label, specify its origin:
     - `"malt"`: derived from an existing profile expertise.
     - `"wikidata"`: taken from Wikidata context.
     - `"gemini"`: a new synthesized composite term.

---

### FORMATTING RULES (CRITICAL)
- **No Internal Quotes:** Do NOT wrap your label string values in single or double quotes.

---

### INPUT DATA

**Target Languages:** {", ".join(languages)}

**Baseline Composite ID:** {qid_combo}
**Is Directional Process:** {is_directional}{directional_hint}

**1. Associated Wikidata Concepts in this compound:**
{wd_text_joined}

**2. Skills in this Group (Use for context to determine majority coherence & missing concepts):**
{malt_text_joined}
\end{lstlisting}
\end{tcolorbox}
\vspace{-2mm}
\captionof{figure}{\footnotesize Canonicalization Prompt - Multiple QID}
\label{fig:canonicalization-compound-prompt}
\vspace{4mm}

\begin{tcolorbox}[academicexample, breakable, enhanced]
\begin{lstlisting}
You are a **Skill Reconciliation Specialist** for orphaned skills.

These skills were REJECTED from their original clusters during curation because they were deemed not equivalent to the cluster's core concept.
Your task is to create COMPLETELY NEW concept identifiers for these orphaned skills.

---

### PRIMARY DIRECTIVES

1. **Fresh Reconciliation**
   - Treat each skill as a potentially NEW concept that needs its own identifier.
   - DO NOT assume these skills fit into their previous Wikidata QIDs.
   - The Wikidata context provided is from their ORIGINAL (failed) matching - use it only as reference to know what they are NOT.

2. **Synthetic ID Creation**
   - Create new synthetic identifiers using the format: `SYNTH_SKILL_XXX` (where XXX is a unique number).
   - If a skill genuinely matches one of the provided Wikidata QIDs, you may use it.
   - If a skill combines multiple concepts, use `&` notation (e.g., `SYNTH_SKILL_001 & Q12345`).
   - Each unique skill concept should get its own unique identifier. Group skills together if they mean the exact same thing.

3. **Preferred Label Generation**
   - For each NEW synthetic ID, create localized preferred labels in all target languages.
   - Set `source` to `"gemini"` for synthesized labels.
   - Set `source` to `"malt"` if you're using a high-frequency Malt term.
   - Set `source` to `"wikidata"` only if you're reusing a Wikidata label.

4. **Skill Validation**
   - Set `is_skill: false` if the text is NOT a valid professional skill.
   - Use `rejection_reason`: `NOT_A_SKILL`, `SEMANTIC_MISMATCH`, or `AMBIGUOUS`.

---

### FORMATTING RULES (CRITICAL)
- **No Internal Quotes:** Do NOT wrap label values in quotes (write `Data Analysis`, NOT `'Data Analysis'`).

---

### INPUT DATA

**Target Languages:** {", ".join(languages)}

**1. Original Wikidata Context (for reference only):**
{wd_text_joined}

**2. Orphaned Skills to Reconcile:**
{malt_text_joined}

Remember: These skills failed to fit their original clusters. Create fresh, appropriate identifiers for them.
\end{lstlisting}
\end{tcolorbox}
\vspace{-2mm}
\captionof{figure}{\footnotesize Canonicalization Prompt - Orphans}
\label{fig:canonicalization-orphan-prompt}
\vspace{4mm}

\section{Curation details}
To maintain structural purity, an active curation agent validates the strict equivalence of every raw skill against its broader canonical grouping. As shown in the first prompt, skills that fail this check are rejected with a specific granular reason (e.g., \texttt{DOMAIN\_SPECIALIZATION} or \texttt{NOT\_A\_SKILL}) and assigned a suggested preferred label, routing them to the 'Orphan' queue for future iteration. A secondary rewriting prompt is then applied to the surviving baseline entity to ensure the final labels are stripped of typographical noise and irrelevant job titles.

\begin{tcolorbox}[academicexample, breakable, enhanced]
\begin{lstlisting}
You are a **Skills Cluster Reviewer** for a freelance marketplace.
Your task is to analyze a list of raw skills and determine if they should be mapped to the established Preferred Names for search indexing.

---

### DIRECTIVES
**MAXIMUM INCLUSION RULE:** We are building a search engine index. You must err on the side of INCLUSION (`is_equivalent: true`). 
A skill is **equivalent** if a client searching for the Preferred Name would reasonably want to hire a freelancer who wrote this raw text.

1. **What to KEEP (`is_equivalent: true`, `rejection_reason: null`):**
   - **Translations (CRITICAL)**: Direct translations of the core concept in English, French, Spanish, German, or Dutch MUST be kept together in the same cluster (e.g., KEEP "Social media" or "Redes sociales" inside the "Réseaux sociaux" cluster).
   - **Synonyms & Paraphrases**: Variations in phrasing.
   - **Typos & Truncations**: Misspellings or cut-off words (e.g., "photosho" for Photoshop, "indes" for InDesign).
   - **Proficiency Modifiers**: (e.g., "Expert in", "Maîtrise de", "Senior").
   - **Role Modifiers & Job Titles**: (e.g., "Consultant SEO", "Freelancer", "Data Analyst", "Project Manager").
   - **Versions**: (e.g., "Python 3" for "Python").

2. **What to REMOVE (`is_equivalent: false`) and Categorize (`rejection_reason`):**
   - `DOMAIN_SPECIALIZATION`: A specific domain or industry that requires distinct technical knowledge. Even if it contains the root word, it must be rejected so it can form its own cluster. (e.g., REMOVE "Web Project Management", "IT Project Management", or "Agile Project Management" from the general "Project Management" cluster).
   - `NOT_A_SKILL`: Completely unrelated to professional work (e.g., "Hard worker", "Available", "I am fast").
   - `SEMANTIC_MISMATCH`: A completely different technical or professional concept.
   - `AMBIGUOUS`: Too vague to map to anything (e.g., "IT", "Consulting", "Management" - when standing alone).
   - `DISTINCT_TOOL_OR_LANGUAGE`: It is a completely different software/language that deserves its own cluster (e.g., removing "Java" from a "JavaScript" cluster).

  ### MULTILINGUAL HANDLING                                                                                                                                                                     
  - Skills may be written in French, Spanish, German, Dutch, or English.      
  - When comparing to Preferred Names, consider cross-language equivalence: 
    - "Gestion de projet" (FR) = "Project Management" (EN)                                 
    - "Projektmanagement" (DE) = "Project Management" (EN) 
  - Proficiency modifiers translate across languages:          
    - "Expert en" (FR), "Experto en" (ES), "Expert in" (EN), "Experte in" (DE) are all equivalent 

### REJECTION PROTOCOL
If 'is_equivalent' is FALSE, you MUST:
1. Select a 'rejection_reason' (e.g., DOMAIN_SPECIALIZATION).
2. When rejecting a skill, you MUST provide a `suggested_pref_label`. This label MUST ALWAYS be written in standard English, regardless of the language of the raw text (e.g., if the raw text is 'audit RH', the suggested label must be 'HR Audit').

---

### FORMATTING RULES
- You MUST evaluate EVERY skill provided in the Input Data.

---

### INPUT DATA
**Preferred Names for Context:**
{json.dumps(pref_labels, indent=2, ensure_ascii=False)}

**Skills to Review:**
{json.dumps(skills, indent=2, ensure_ascii=False)}
\end{lstlisting}
\end{tcolorbox}
\vspace{-2mm}
\captionof{figure}{\footnotesize Curation Prompt}
\label{fig:curation-validation-prompt}
\vspace{4mm}

\begin{tcolorbox}[academicexample, breakable, enhanced]
\begin{lstlisting}
You are an expert **Multilingual Taxonomy Linguist and HR Specialist**.
Review and refine the current "Preferred Labels" for professional skill clusters across these languages: {languages}.

### DIRECTIVES
Act as a local HR specialist in the country of the target language.
- **Professional Context:** Ignore literal dictionary translations. Use standard CV/Job Description terms (e.g., "Consulting" instead of "Conseil" in French).
- **Anglicisms:** If the English term is the dominant standard (common in Tech/Business), keep it! (e.g., "Cloud Computing" for German).
- **Grammar Standard:** Use Singular Nouns or Gerunds representing the capability, not the action (e.g., "Management" instead of "To Manage").

### PURIFICATION RULE (CRITICAL)
The "Alternate Skills" list provided below contains messy search data. It includes job titles (e.g., "Consultant SEO"), seniority levels (e.g., "Expert"), and typos. 
**Your generated labels MUST STRIP OUT all of this noise.** Do not include words like "Consultant", "Expert", or "Freelance" in your final labels. Extract ONLY the pure, canonical underlying skill.

### INPUT DATA
**Current Preferred Labels:**
{json.dumps(pref_labels, indent=2)}

**Alternate Skills in Cluster (for context):**
{json.dumps(alternate_skills, indent=2)}
\end{lstlisting}
\end{tcolorbox}
\vspace{-2mm}
\captionof{figure}{\footnotesize Rewriting Prompt}
\label{fig:curation-rewriting-prompt}
\vspace{4mm}

\section{Consolidation details}
To ensure global structural consistency across incremental batch runs, the consolidation phase evaluates candidate pairs flagged by lightweight lexical heuristics to merge overlapping sub-graphs. The prompt below instructs the model to determine whether to merge or keep entities separate based on semantic overlap, multilingual equivalency, and versioning constraints, outputting a definitive decision and a surviving composite ID.

\begin{tcolorbox}[academicexample, breakable, enhanced]
\begin{lstlisting}
You are an expert **Taxonomy Architect**. 
Two skill clusters have been flagged as potential duplicates. Your goal is to determine if they represent the exact same professional concept.

### DECISION CRITERIA
**MERGE** if ANY of these conditions apply:
- **Exact Synonyms:** "NodeJS" and "Node.js", "React" and "ReactJS"
- **Typo Variants:** "Javscript" and "JavaScript", "Managment" and "Management"
- **Multilingual Equivalents:** "Gestion de projet" (FR) and "Project Management" (EN) with same scope
- **Notation Differences:** "C++" and "C plus plus", "C#" and "C Sharp"

**KEEP_SEPARATE** if ANY of these conditions apply:
- **Specialization vs General:** "Technical Project Management" vs "Project Management" (one is narrower)
- **Version Requiring Different Expertise:** "Angular 1.x" vs "Angular 2+" (fundamentally different frameworks)
- **Completely Different Domains:** "Java" (language) vs "JavaScript" (unrelated language)
- **Methodology vs Tool:** "Agile Project Management" vs "Project Management" (one adds methodology)
- **Context Specificity:** "Remote Project Management" vs "Project Management" (one adds context)

### MULTILINGUAL VALIDATION
- If clusters have different primary labels but same QIDs, they're likely multilingual variants then **MERGE**
- If clusters have similar English labels but different validated skills, they're likely distinct then **KEEP_SEPARATE**

### SURVIVOR RULES
If you MERGE, you must pick the best ID to keep:
1. **Official over Synthetic:** Prioritize Wikidata QIDs (e.g., Q123) over Orphan IDs (e.g., ORPHAN_ITER1_...).
2. **Lower QID Number:** If both are Wikidata QIDs, prefer the one with the lower number (older, more established broader concept).
3. **Stability:** If merging an Orphan into a QID, the QID MUST be the `surviving_composite_id`.

### INPUT DATA

**CLUSTER A:**
- ID: {pair_data['cluster_1_id']}
- Refined Labels (Filtered):
{cluster_1_labels}
- Validated Skills in this Cluster: {json.dumps(pair_data['cluster_1_skills'], ensure_ascii=False)}

**CLUSTER B:**
- ID: {pair_data['cluster_2_id']}
- Refined Labels (Filtered):
{cluster_2_labels}
- Validated Skills in this Cluster: {json.dumps(pair_data['cluster_2_skills'], ensure_ascii=False)}

---
Based on the labels and skills provided, decide if these two clusters should be merged.
\end{lstlisting}
\end{tcolorbox}
\vspace{-2mm}
\captionof{figure}{\footnotesize Consolidation Prompt}
\label{fig:consolidation-prompt}
\vspace{4mm}

\end{document}